\documentclass{article}
\usepackage{graphicx}
\usepackage{caption}
\usepackage{subcaption}

\usepackage[final]{nips_2017}

\usepackage[utf8]{inputenc} 
\usepackage[T1]{fontenc}    
\usepackage{hyperref}       
\usepackage{url}            
\usepackage{booktabs}       
\usepackage{amsfonts}       
\usepackage{nicefrac}       
\usepackage{microtype}      

\title{Deep Learning for Identifying Potential Conceptual Shifts for Co-creative Drawing}

\author{
Pegah Karimi\\
University of North Carolina at Charlotte\\
Charlotte, NC 28223 USA\\
\texttt{pkarimi@uncc.edu} \\
\And
Nicholas Davis\\
University of North Carolina at Charlotte\\
Charlotte, NC 28223 USA\\
\texttt{ndavis64@uncc.edu} \\
\And
Kazjon Grace\\
The University of Sydney\\
Sydney, NSW 2006 Australia\\
\texttt{kazjon.grace@sydney.edu.au} \\
\And
Mary Lou Maher\\
University of North Carolina at Charlotte\\
Charlotte, NC 28223 USA\\
\texttt{m.maher@uncc.edu}\\
}

\begin{document}
\maketitle

\begin{abstract}
We present a system for identifying conceptual shifts between visual categories, which will form the basis for a co-creative drawing system to help users draw more creative sketches. The system recognizes human sketches and matches them to structurally similar sketches from categories to which they do not belong. This would allow a co-creative drawing system to produce an ambiguous sketch that blends features from both categories.

\end{abstract}

\section{Introduction}

Creative sketching is important in a wide variety of domains including engineering, art, and architecture [1]. We propose a co-creative system that plays a free association drawing game that facilitates more creative sketches by introducing conceptual shifts. In co-creative systems, an AI and a user collaborate on a creative task [2]. Our proposed AI agent recognizes objects in human sketches and responds by suggesting a conceptual shift. In this context, a conceptual shift entails recognizing an object from a different conceptual category with which the current object shares structural characteristics. The goal is to respond with another sketch that leverages this conceptual shift, such as by drawing a conceptual blend [3] of the two objects. The resulting free association game would encourage the fluent expression of ideas by the user. The system described here forms the basis of a co-creative tool that helps the user to learn creative sketching.
 
\section{Training the co-creative agent with sketches}

We use clustering on deep features to train the co-creative agent to generate a visual representation of sketches that enable a conceptual shift. This training has two steps:

\textbf{Learning the visual representation of sketches.} Deep learning architectures are able to extract highly detailed visual representations from images leading to high classification accuracy and high-quality generated images [4]. We use the Google Quick, Draw! dataset [5] in place of user input and train our system on a subset of 65 categories with up to 110,000 images in each. We represent each sketch as a high-level feature vector obtained using VGG-16 [6], a standard convolutional neural network (CNN) architecture. We started with a model that was pre-trained on the ImageNet dataset [7] and then fine-tuned the weights by training on the sketches. The model contains 15 convolutional layers and two fully connected layers. We extract the features from the first fully connected layer which produces 4096 features per image, and use these features to represent each sketch.

\begin{figure}[h]
\centering
\includegraphics[width=0.6\textwidth]{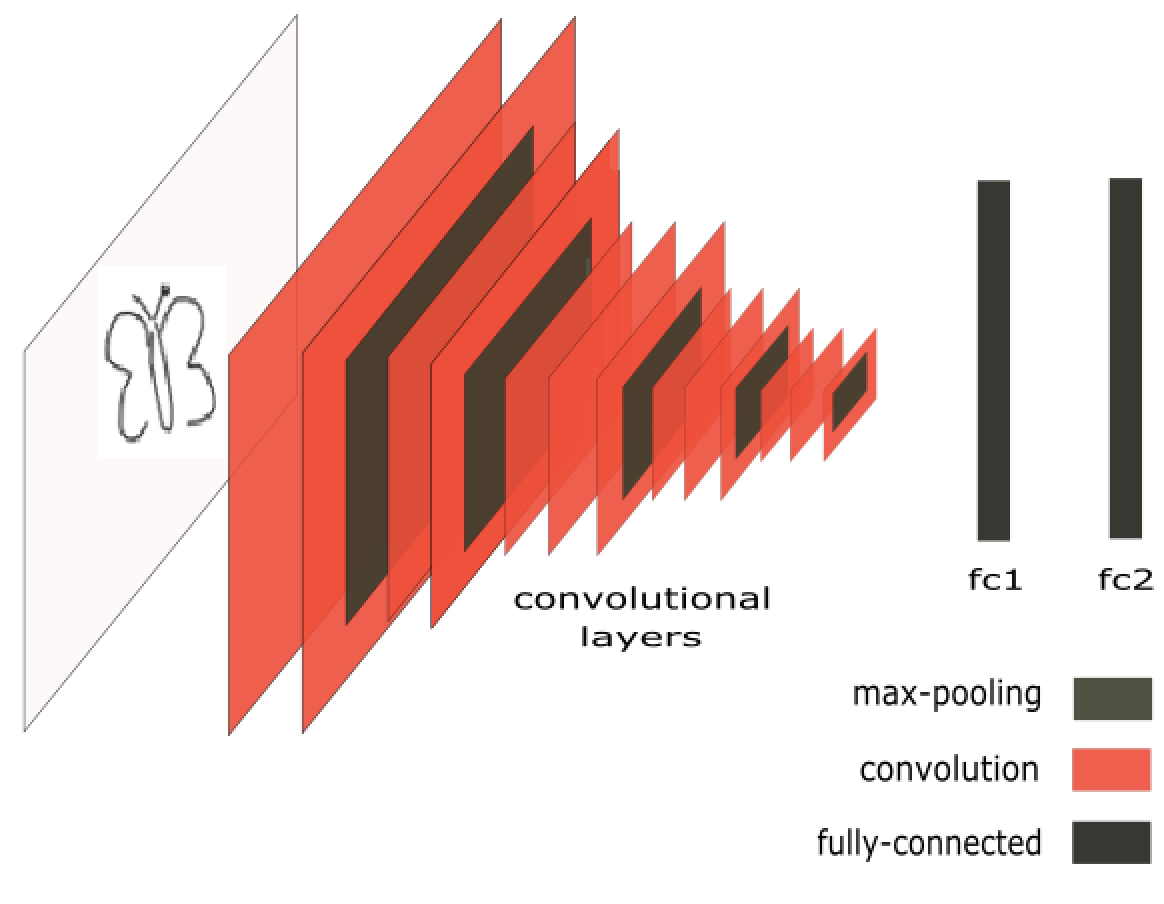}
\caption{The neural representation of sketches, drawn from the activation of the first fully connected layer (fc1) in the VGG-16 Convolutional Neural Network architecture [6].} 
\end{figure}

\textbf{Clustering sketches into sets.}  As the objects of each category have high variance (e.g. different shape, viewpoint and angle), we separate each category using clustering on the VGG-16 representation, which identifies structurally similar sub-categories that we use for identifying potential conceptual shifts. This is due to the fact that humans have different perceptions about drawing the stereotypical shape of an object, which leads to highly different images despite belonging to the same category. We used k-means algorithm and selected k via the elbow method.  

\begin{figure}[h]
	\begin{subfigure}[t]{2.5in}
		\centering
		\includegraphics[width=2.5in]{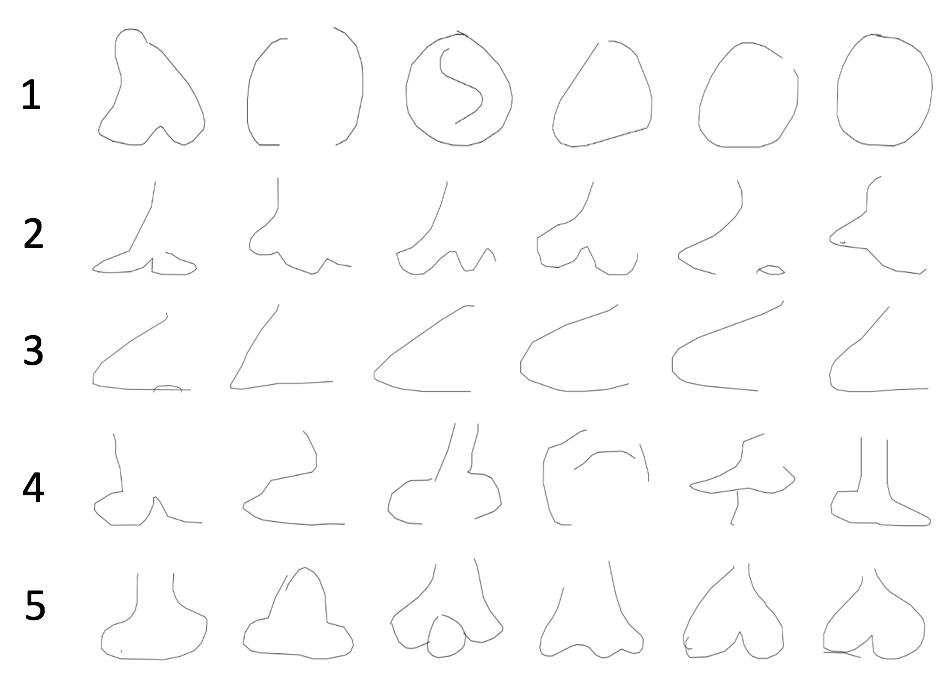}
		\caption{Samples of five clusters from nose category}\label{fig:five_nose}	
	\end{subfigure}
	\hspace{0.2in}
	\begin{subfigure}[t]{2.35in}
		\centering
		\includegraphics[width=2.40in,height=1.8in]{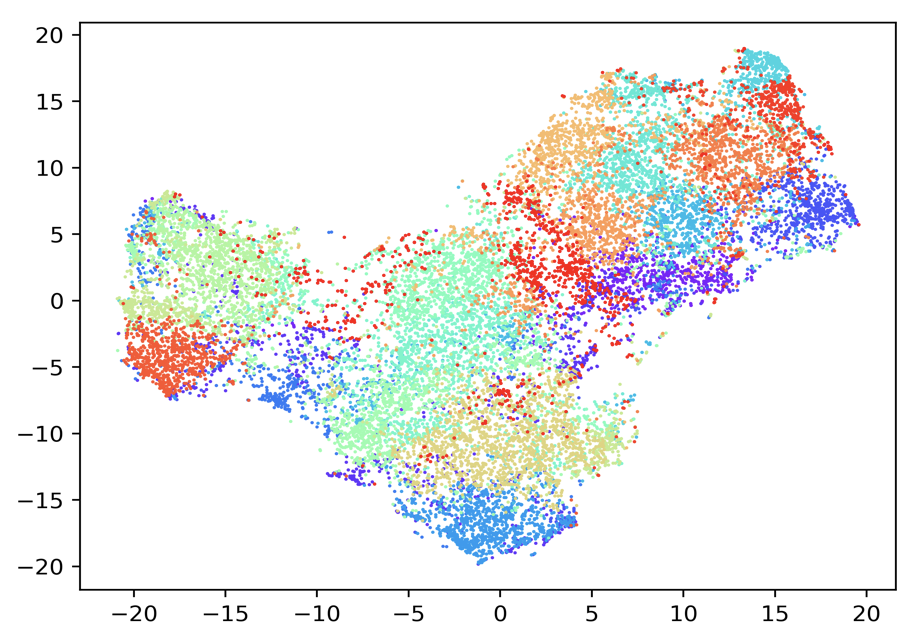}
		\caption{embedding visualization of nose category}\label{fig:embed_nose}
	\end{subfigure}
\caption{Clustering samples (a) and visualization using LargeVis [8] (b).}
\end{figure}

\section{The Co-creative System for drawing}

In our proposed co-creative drawing system a user and an agent take turns to sketch. The agent has to perceive the user input and identify opportunities for conceptual shifts. To identify potential conceptual shifts we match objects from one category to structurally similar objects in another category. We compare the Euclidean distances between the centroids of the cluster (i.e. sub-category) to which the sketch belongs and clusters from other categories. We identify the cluster with the minimum distance to the current object's cluster and use this to propose a conceptual shift.  Our approach involves 3 steps: 

\textbf{Recognizing.} The agent takes the user’s sketch (currently a randomly selected sketch from the dataset), extracts its feature vector, and determines its cluster.

\textbf{Matching.} The user’s sketch is matched to the most similar cluster that isn’t in the same object category as the current sketch.

\textbf{Contributing.} The agent contributes (currently by random selection from the dataset, but in future by neural drawing) a sketch from the cluster identified as a potential conceptual shift.

\section{Case Study Demonstrating our Conceptual Shift Algorithm}
To demonstrate the utility of our algorithms, we provide a case study exploring how the system could respond to users’ sketches. The case study explores the different ways the system could respond to the same input sketch.

\textbf{Call and Response with Conceptual Shifts.} In this scenario, a user’s sketch is matched to a sketch from another category. The algorithm determines which category is most structurally similar to the user’s input and then generates a response based on that calculation. Figure 3 and Figure 4 show the resulting proposed conceptual shifts from 6 input categories in each. The Euclidean distance is used to determine which category is the closest to each input sketch. Figure 3 shows input-response pairs, those that are very similar (i.e. have a low Euclidean distance, signifying a good potential conceptual shift), and Figure 4 shows those for which the closest match was less similar (i.e. a higher distance, perhaps signifying less potential for a conceptual shift).

\begin{figure}[h]
\centering
\includegraphics[width=\textwidth,height=2.29in]{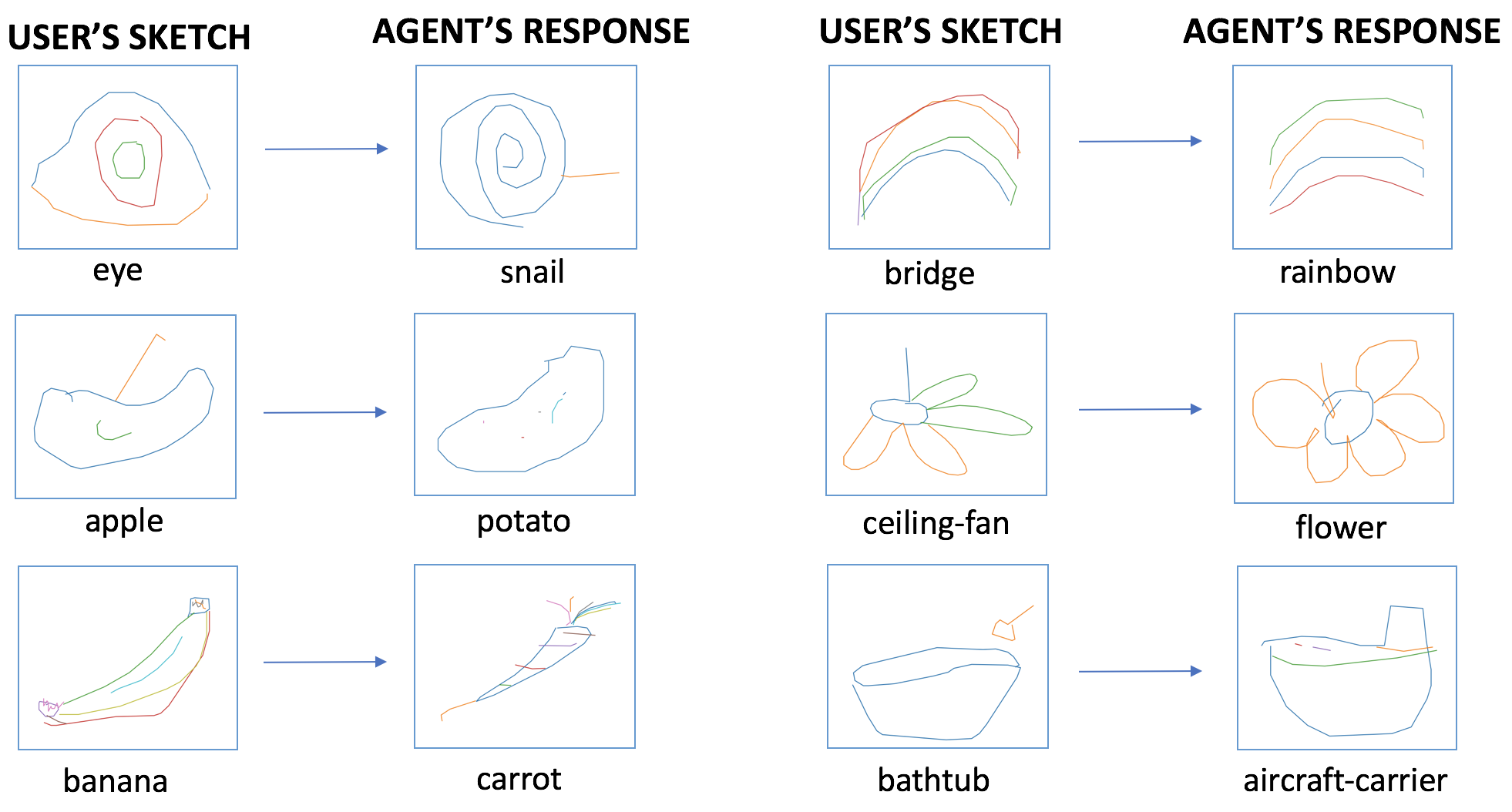}
\caption{Examples from six categories with conceptual shifts that are very visually similar to the input (lower distance between matched clusters). Each color represents a different stroke.} 
\end{figure}

\begin{figure}[h]
\centering
\includegraphics[width=\textwidth,height=2.29in]{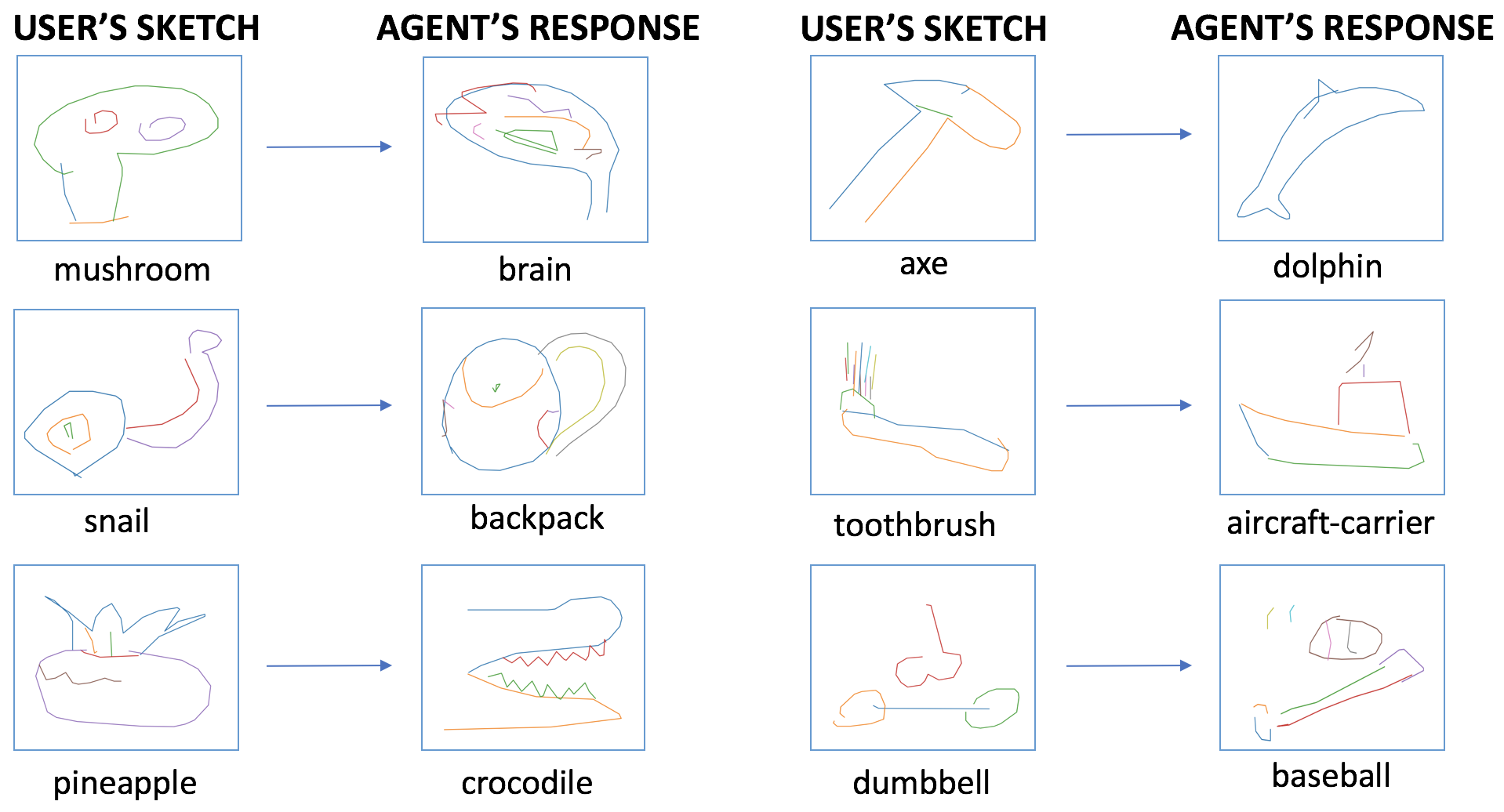}
\caption{Examples from six categories with conceptual shifts that are less visually similar to the input (higher distance between matched clusters).} 
\end{figure}

\textbf{Single Input and Multiple Responses with Conceptual Shifts.} In the second scenario of our case study (Figure 5 and Figure 6), the user draws one input item and the system generates many potential conceptual shifts from different categories. The user began by sketching an object. After the system extracted the visual features from the object, it matched five potential conceptual shifts. 

\begin{figure}[ht]
\centering
\includegraphics[width=1.0\textwidth,height=2in]{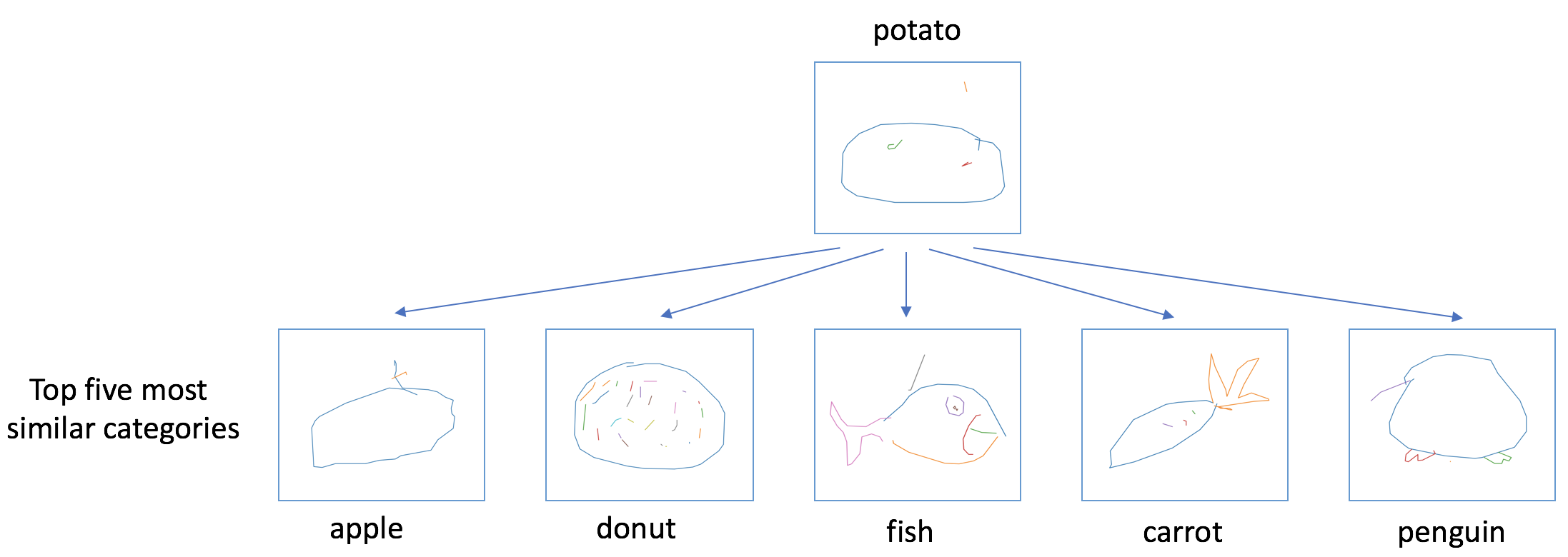}
\caption{Top 5 potential conceptual shift responses to a single sketch of a potato.}
\end{figure}

\begin{figure}[ht]
\centering
\includegraphics[width=1.02\textwidth,height=1.9in]{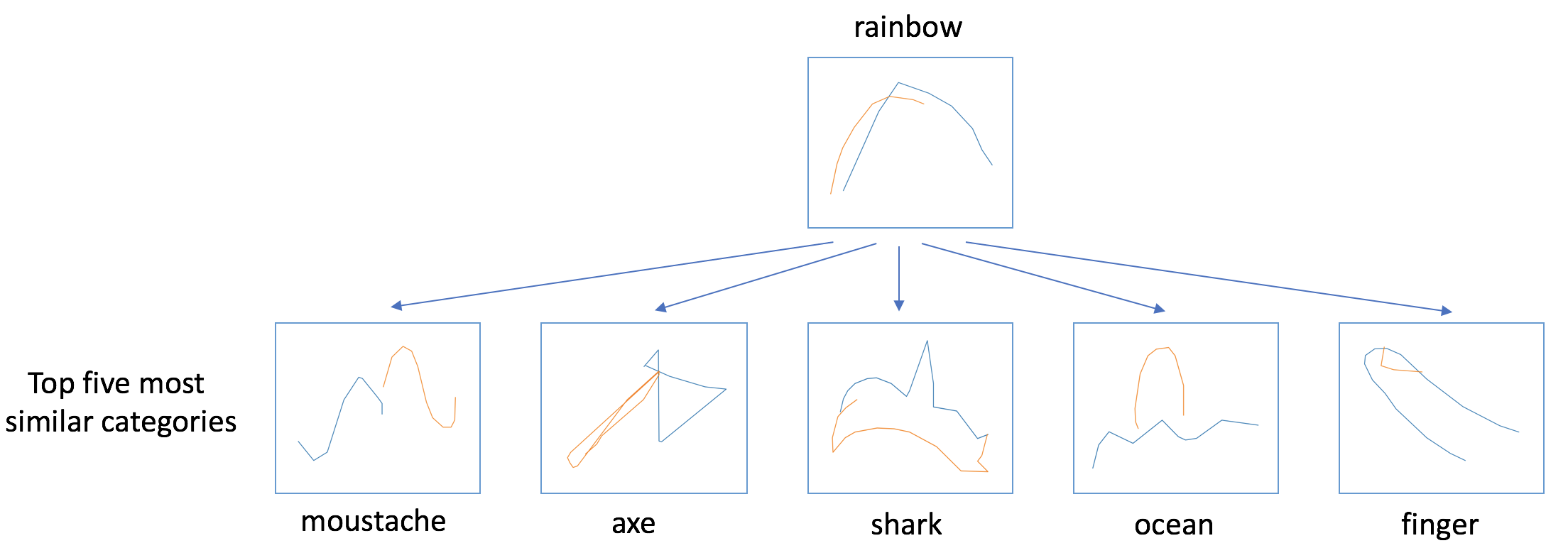}
\caption{Top 5 potential conceptual shift responses to a single sketch of a rainbow.}
\end{figure}

\textbf{Multiple Inputs and Multiple Responses with Conceptual Shifts.} In the final scenario, shown in Figure 7 and Figure 8, the user produces multiple iterations of an object in the same category. The system selects a different conceptual shift category for each input. In this case, the user drew several types of an object that each have unique structural characteristics. The algorithm finds different potential conceptual shifts in each case.

\begin{figure}[ht]
\centering
\includegraphics[width=1.02\textwidth,height=2.0in]{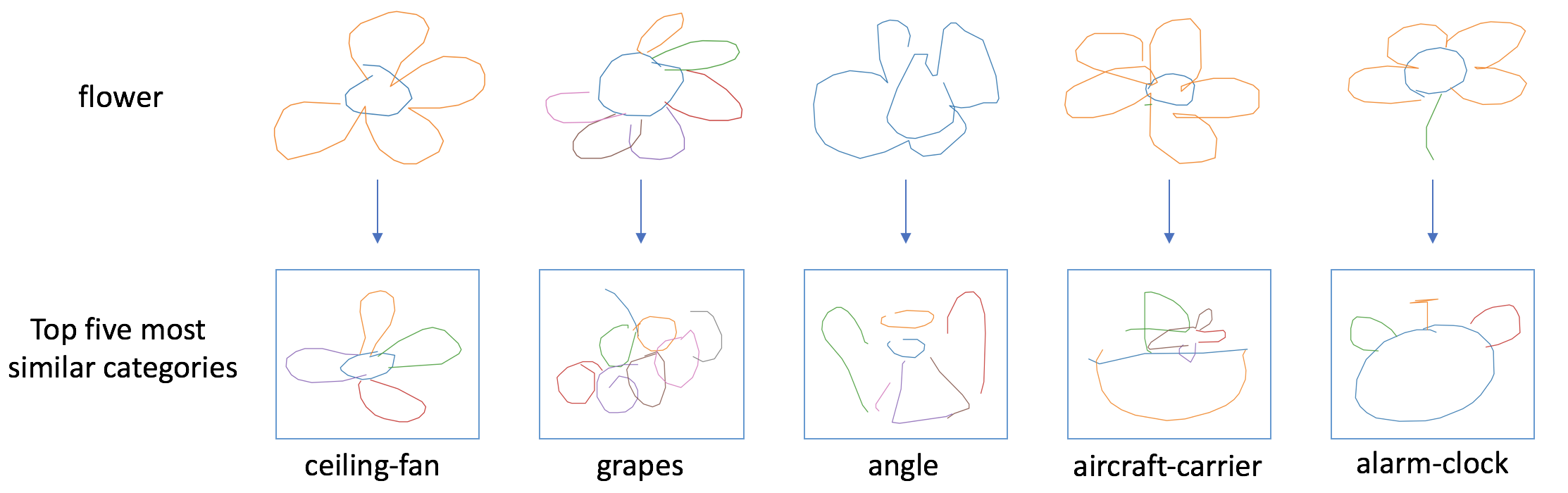}
\caption{The system’s conceptual shift responses to multiple flower sketches.}
\end{figure}

\begin{figure}[ht]
\centering
\includegraphics[width=1.02\textwidth,height=2.0in]{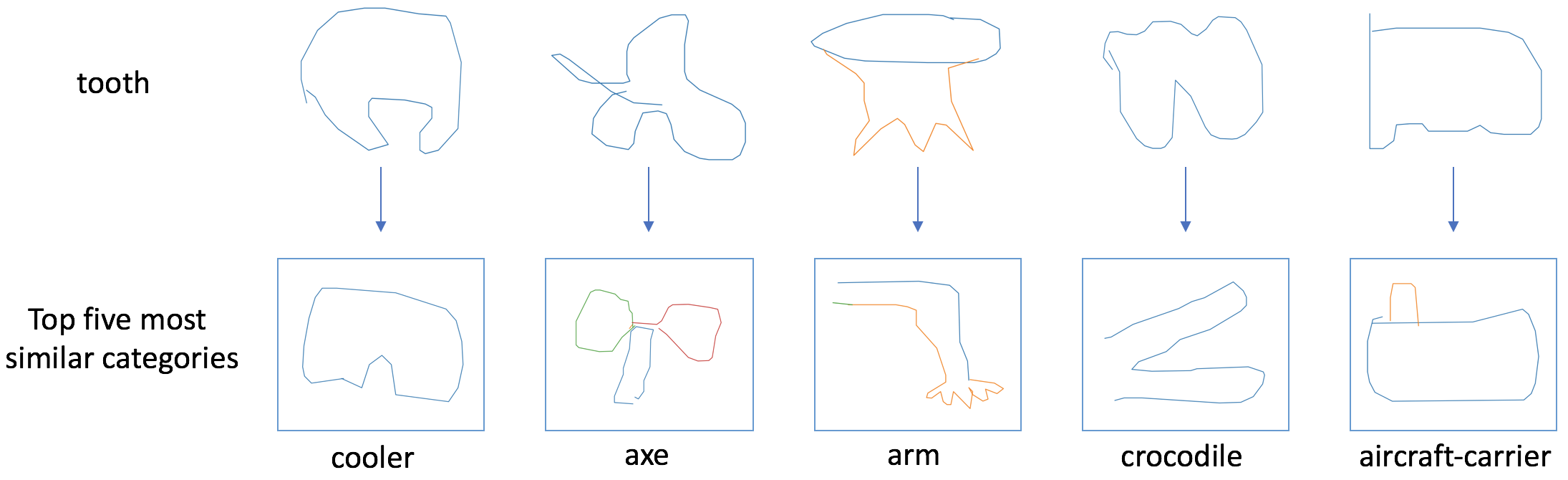}
\caption{The system’s conceptual shift responses to multiple tooth sketches.}
\end{figure}

\section{Conclusions and Future Work}

In this paper we described a co-creative drawing system that plays a free association game. We described a framework that identifies a conceptual shift through recognizing, matching and contributing. This system will be used as the basis for a drawing agent that creates conceptual blends that encourage user creativity. In the future we will investigate alternative architectures for the encoder to improve the sketch representation for clustering. In addition, we would like to develop a more comprehensive matching algorithm for conceptual shifts, and incorporate a neural approach to drawing.

\section*{References}

\medskip
\small

[1] Ullman, D. G., Wood, S.\ \& Craig, D.\ (1990). The importance of drawing in the mechanical design process. {\it Computers and graphics}, 14(2), 263--274.

[2] Davis, N., Hsiao, C. P., Yashraj Singh, K., Li, L.\ \& Magerko, B.\ (2016, March). Empirically studying participatory sense-making in abstract drawing with a co-creative cognitive agent. In {\it Proceedings of the 21st International Conference on Intelligent User Interfaces} (pp. 196-207). ACM.

[3] Fauconnier, G.\ \& Turner, M.\ (2008). {\it The way we think: Conceptual blending and the mind's hidden complexities.} Basic Books.

[4] LeCun, Y., Bengio, Y.\ \& Hinton, G.\ (2015). Deep learning. {\it Nature}, 521(7553), 436-444.

[5] Ha, D.\ \& Eck, D.\ (2017). A Neural Representation of Sketch Drawings. {\it arXiv preprint arXiv:1704.03477.}

[6] Simonyan, K.\ \& Zisserman, A.\ (2014). Very deep convolutional networks for large-scale image recognition. {\it arXiv preprint arXiv:1409.1556.}

[7] Deng, J., Dong, W., Socher, R., Li, L. J., Li, K.\ \& Fei-Fei, L.\ (2009, June). Imagenet: A large-scale hierarchical image database. In {\it Computer Vision and Pattern Recognition, 2009. CVPR 2009. IEEE Conference on} (pp. 248-255). IEEE.

[8] Tang, J., Liu, J., Zhang, M.\ \& Mei, Q.\ (2016, April). Visualizing large-scale and high-dimensional data. In {\it Proceedings of the 25th International Conference on World Wide Web} (pp. 287-297). International World Wide Web Conferences Steering Committee.

\end{document}